\documentclass[conference]{IEEEtran}
\IEEEoverridecommandlockouts
\usepackage{cite}
\usepackage{amsmath,amssymb,amsfonts}
\usepackage{algorithmic}
\usepackage{graphicx}
\usepackage{textcomp}
\usepackage{xcolor}

\usepackage{subfigure}
\usepackage{bbm}
\usepackage{multirow}
\usepackage{booktabs}
\usepackage{hyperref}

\def\BibTeX{{\rm B\kern-.05em{\sc i\kern-.025em b}\kern-.08em
    T\kern-.1667em\lower.7ex\hbox{E}\kern-.125emX}}

\begin{document}

\title{Model Discrepancy Learning: Synthetic Faces Detection Based on Multi-Reconstruction}

\author{
    QingChao Jiang$^{1}$, 
    ZhiShuo Xu$^{1}$, 
    ZhiYing Zhu$^{1}$\textsuperscript{*}\thanks{*Our dataset at https://github.com/Hurrice-star/ASFD.}, 
    Ning Chen$^{1}$, 
    HaoYue Wang$^{2}$, 
    ZhongJie Ba$^{3}$ \\
    \emph{$^{1}$ East China University of Science and Technology, Shanghai, China} \\
    \emph{$^{2}$ Fudan University, Shanghai, China} \\
    \emph{$^{3}$ The State Key Laboratory of Blockchain and Data Security, Zhejiang University, Hangzhou, China} 
}

\maketitle

\begin{abstract}
Advances in image generation enable hyper-realistic synthetic faces but also pose risks, thus making synthetic face detection crucial. Previous research focuses on the general differences between generated images and real images, often overlooking the discrepancies among various generative techniques. In this paper, we explore the intrinsic relationship between synthetic images and their corresponding generation technologies. We find that specific images exhibit significant reconstruction discrepancies across different generative methods and that matching generation techniques provide more accurate reconstructions. Based on this insight, we propose a Multi-Reconstruction-based detector. By reversing and reconstructing images using multiple generative models, we analyze the reconstruction differences among real, GAN-generated, and DM-generated images to facilitate effective differentiation. Additionally, we introduce the Asian Synthetic Face Dataset (ASFD), containing synthetic Asian faces generated with various GANs and DMs. This dataset complements existing synthetic face datasets. Experimental results demonstrate that our detector achieves exceptional performance, with strong generalization and robustness.

\end{abstract}

\begin{IEEEkeywords}
synthetic face detection, reconstruction discrepancies, ASFD
\end{IEEEkeywords}


\section{Introduction}
The rapid advancement of generative models has made distinguishing between real and synthetic images increasingly challenging. Synthetic human faces, created by models like Generative Adversarial Networks (GANs) \cite{goodfellow2020generative} and diffusion models (DMs) \cite{sohl2015deep}, are widely used in fields such as entertainment, virtual reality, and data augmentation. However, these highly realistic synthetic faces are also exploited for malicious purposes, such as identity theft, misinformation, and automated deception, raising significant security concerns. Synthetic face detection has become crucial.

Existing studies treat synthetic image detection as a binary classification task, distinguishing between real and synthetic images. Some methods \cite{barni2020cnn,wang2020cnn,tan2023learning} rely on data-driven binary classifiers. These classifiers overlook the intrinsic discrepancies between GAN-generated and DM-generated images. As a result, they often suffer from poor generalization and performance. Other researchers \cite{barni2020cnn,frank2020leveraging,wang2020cnn,tan2023learning,wang2023dire,ricker2024aeroblade} design detection models tailored to specific generative models, effectively leveraging their unique features. While these models excel at recognizing known synthetic types, they struggle in mixed scenarios (i.e., distinguishing between GAN-generated and DM-generated images when both are present) due to limited generalization and robustness. Recent studies \cite{pontorno2024deepfeaturex,guarnera2024mastering} explore a three-class classification task, which distinguishes real, GAN-generated, and DM-generated images. They can better utilize the information embedded in the data distribution, thereby improving detection performance. However, these methods remain heavily data-driven and fail to leverage the distinctive attributes of GANs and DMs effectively. In practical applications, synthetic face detection models encounter a limited variety of racial types while requiring high accuracy. Existing datasets are racially diverse but unbalanced. For example, the FFHQ \cite{karras2019style} dataset includes only a small number of Asian faces compared to a large number of European faces. This imbalance somewhat restricts research on specific racial groups.

\begin{figure}[t]
\centering
\includegraphics[width=0.98\linewidth]{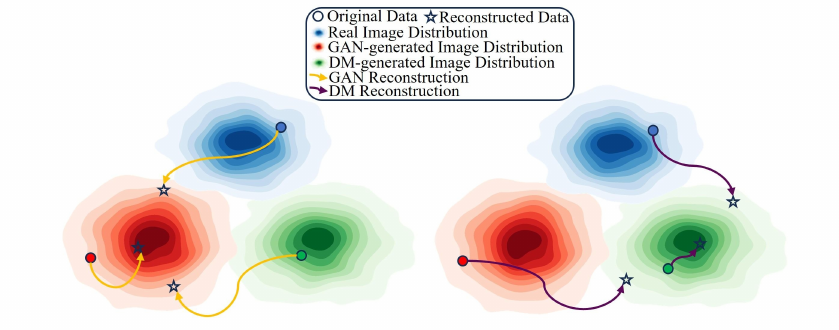}
\caption{GANs and DMs achieve higher reconstruction quality for data aligned with their distributions, while the quality degrades for data outside these distributions. Specifically, well-reconstructed images are near the center of the distribution, whereas poorly reconstructed ones lie at the periphery.}
\label{fig:1a}
\end{figure}

In this paper, we explore and perform discrepancy analysis on the intrinsic relationship between synthetic images and the generative techniques that produce them. As shown in Fig. \ref{fig:1a}, generative models achieve the best reconstruction performance on images they have generated, while exhibiting inferior performance when reconstructing images from other sources. Based on this observation, we propose a Multi-Reconstruction-based Detector, which leverages GAN and DM to invert and reconstruct input images. By capturing the subtle differences in reconstruction performance, our method effectively distinguishes between real, GAN-generated, and DM-generated images. Furthermore, to address the underrepresentation of Asian face datasets, we introduce a novel dataset called the Asian Synthetic Face Dataset (ASFD). This dataset includes synthetic Asian faces generated by four classic GANs \cite{karras2018progressive,karras2019style,karras2020analyzing,esser2021taming} and four DMs \cite{dhariwal2021diffusion,nichol2021improved,rombach2022high,song2020score}. It could provide valuable support and serve as a reference for tasks targeting Asian populations.

Our contributions can be summarized as follows:
\begin{itemize}
\item We propose the Multi-Reconstruction-based Detector to address the challenging task of distinguishing between real, GAN-generated, and DM-generated faces.
\item We introduce the ASFD dataset to address the underrepresentation of Asian synthetic face data and provide valuable support and a reference for tasks targeting Asian populations.
\item Extensive experiments demonstrate that our method significantly improves detection performance and robustness across different generative models.  
\end{itemize}


\section{Related Work}

\subsection{Image Generation}
The two most mainstream approaches to image generation today are GAN \cite{goodfellow2020generative} -generated and DM \cite{sohl2015deep} -generated techniques, which are widely applied in areas such as synthetic face generation. GANs \cite{goodfellow2020generative} consist of a generator and a discriminator. The generator produces synthetic images, while the discriminator evaluates the authenticity of the images. Through adversarial training between the generator and the discriminator, GANs enable the generator to produce high-quality images, achieving remarkable results in applications such as face generation.

DMs \cite{sohl2015deep} generate images through a probabilistic process that involves adding noise to an image and then reversing the process to denoise it step by step. Compared to GANs \cite{goodfellow2020generative}, which excel at generating high-resolution images, DMs \cite{sohl2015deep} have the advantage of producing more consistent results and better covering complex data distributions. In recent years, DMs \cite{sohl2015deep} have been widely adopted, including models such as ADM \cite{dhariwal2021diffusion}, LDM \cite{rombach2022high}, and SDE \cite{song2020score}. These models excel in controlling various aspects of the generation process, enabling the production of highly realistic and detailed images.

\subsection{Synthetic image detection}
Synthetic image detection seeks to distinguish whether an image is real (captured directly by cameras or smartphones) or synthetic (generated by generative models). Early approaches often relied on data-driven binary classifiers. For example, Reference \cite{barni2020cnn} proposed a CNN-based detector, which demonstrated strong performance in detecting synthetic images under closed-set conditions. Reference \cite{wang2020cnn} advanced the field by demonstrating that a simple ResNet-50-based classifier, trained on images generated by ProGAN, could effectively generalize to unseen GAN-generated images. Reference \cite{tan2023learning} proposed extracting gradient maps from a well-trained image classifier as image fingerprints and performing binary classification based on these maps. These approaches overlook the discrepancies between GAN and DM, leading to limited performance in mixed scenarios.

Subsequent research focused on detection models tailored to specific generative models. For example, Reference \cite{frank2020leveraging} analyzed the frequency domain and observed consistent abnormal patterns in synthetic images, leading to a frequency-based detector for GAN-generated images. Reference \cite{wang2023dire} proposed that the residuals from DDIM image reconstruction can be used to predict the authenticity of DM-generated images. Reference \cite{ricker2024aeroblade} proposed a training-free approach for detecting DM-generated images. The method utilizes an autoencoder (AE) to compute reconstruction errors, followed by the LPIPS metric \cite{zhang2018unreasonable} to quantify the similarity between original and reconstructed images. The limitation of these approaches lies in their reliance on specific types of knowledge, which hinders generalization in mixed scenarios.

Recent studies \cite{pontorno2024deepfeaturex,guarnera2024mastering} explore a three-class classification task, which distinguishes real, GAN-generated, and DM-generated images. They can better utilize the information embedded in the data distribution, thereby improving detection performance. However, they rely solely on data-driven methods and fail to account for the fundamental differences between GAN and DM. As a result, their generalization and robustness are limited.

\begin{figure}[t]
\centering
\includegraphics[width=0.85\linewidth]{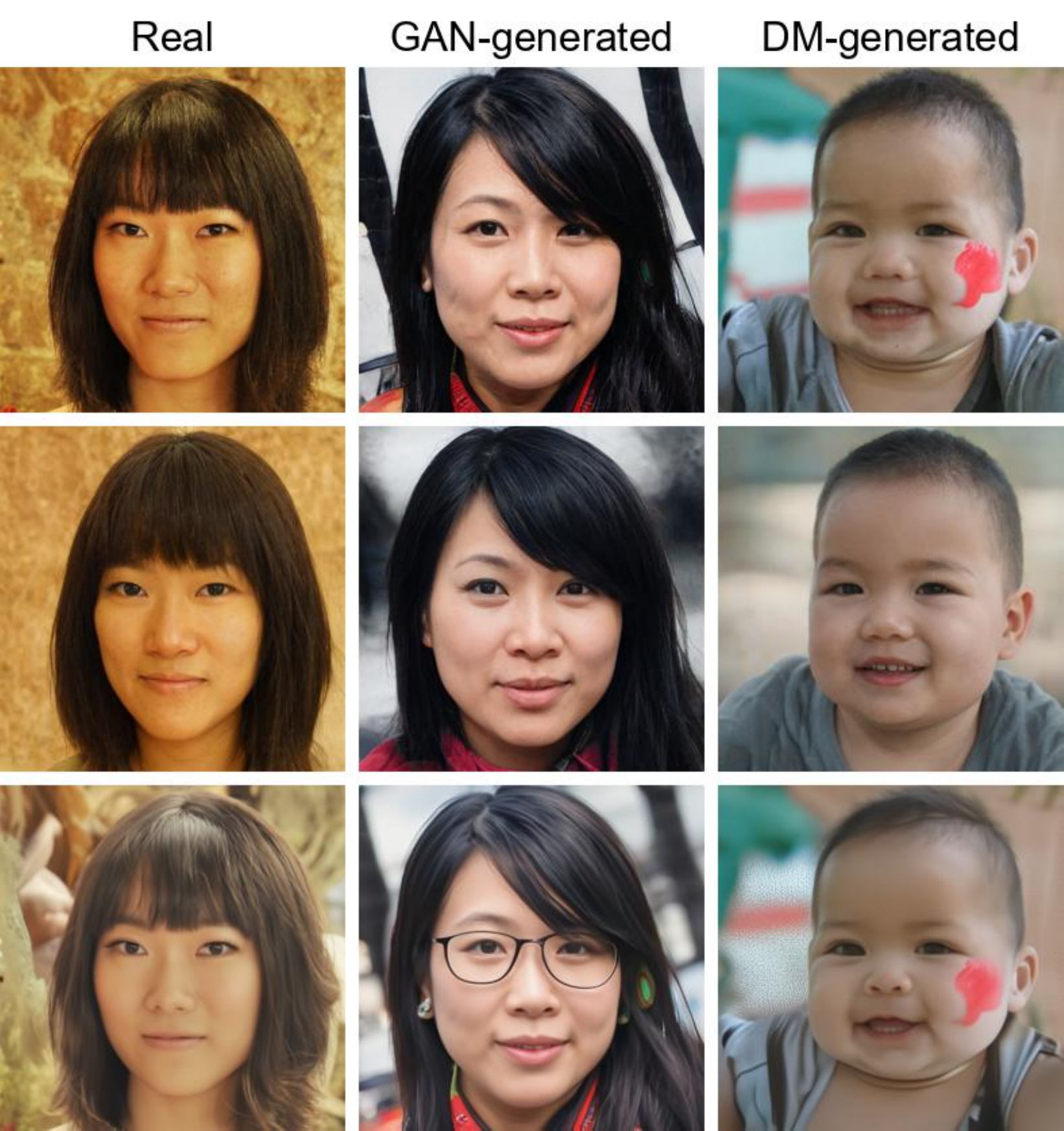}
\caption{Reconstructed images of different categories of images. The first row shows the original images, the second row displays the images reconstructed by GAN, and the third row presents the images reconstructed by DM.}
\label{fig:3}
\end{figure}

\begin{figure*}
\centering
\includegraphics[width=1\linewidth]{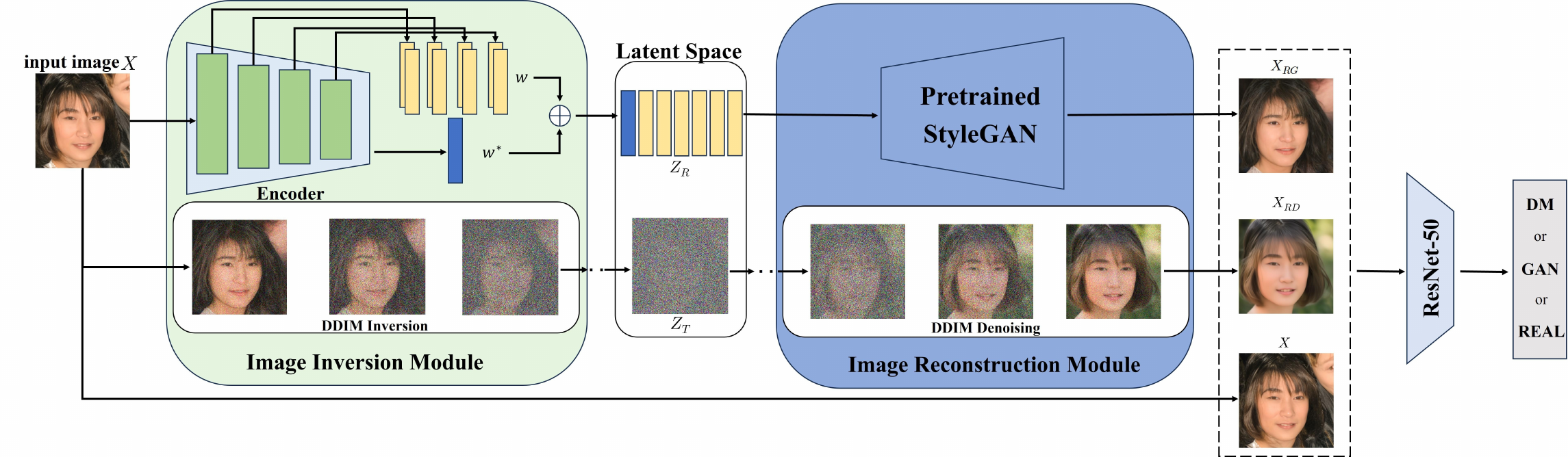}
\caption{\textbf{Multi-Reconstruction-based Detector.} Given an input image $X$, the encoder extracts hierarchical latent codes $w$ and feature codes $w^{*}$, which are concatenated to form the latent representation $Z_{R}$ in the GAN's latent space. This representation is fed into a pre-trained StyleGAN to obtain the reconstructed image $X_{RG}$. Simultaneously, the input image is inverted to the latent representation $Z_{T}$ in a DM's latent space using DDIM inversion. The reconstructed image $X_{RD}$ is then obtained through the DDIM denoising process. Finally, $X$, $X_{RG}$ and $X_{RD}$ are input into a ResNet-50 network for further processing.}
\label{fig:2}
\end{figure*}


\section{Proposed Method}
In this section, we first present the motivation and rationale based on our findings. Next, we introduce the proposed Multi-Reconstruction-based Detector. Finally, we provide a detailed description of the ASFD dataset.

\subsection{Motivation and Intuition}
Studies such as DIRE \cite{wang2023dire} demonstrate that compared to real images, synthetic images generated by DMs can be reconstructed with greater precision using DMs. Inspired by this observation, we aim to generalize this characteristic to other generative techniques. We hypothesize that synthetic images can be more accurately reconstructed by the generative techniques that created them. To validate this hypothesis, we use StyleGAN \cite{karras2019style} and ADM \cite{dhariwal2021diffusion} to invert and reconstruct real, GAN-generated, and DM-generated images. Fig. \ref{fig:3} provides examples of reconstructed images under various scenarios. For real images, both GAN and DM reconstructions exhibit varying degrees of quality degradation and content differences. In contrast, GAN-generated or DM-generated images are precisely reconstructed by their corresponding generative models, while heterogeneous models often alter the original image to some extent. During reconstruction, GANs tend to introduce artifacts, such as unnatural textures or over-smoothed areas. On the other hand, DMs typically preserve finer details but may struggle with certain structures, such as sharp edges or complex textures. Based on this insight, we designed and implemented the Multi-Reconstruction-based Detector to identify and exploit these inversion-reconstruction discrepancies for detecting synthetic images.

\subsection{Multi-Reconstruction-based Detector}
Our Multi-Reconstruction-based Detector, as shown in Fig. \ref{fig:2}. Given an input image, we employ a multi-inversion-reconstruction strategy, combining GAN and DM inversion and reconstruction to generate reconstructed images. Unlike the methods described in \cite{wang2023dire} using reconstruction residuals as inputs, our approach combines the reconstructed images with the original input and feeds them into a convolutional neural network to distinguish between DM-generated, GAN-generated, and real images.

\subsubsection{GAN Inversion and Reconstruction}
To reconstruct images using a GAN, GAN inversion is required. GAN inversion refers to mapping a given image to the latent space of a GAN, enabling the GAN to generate an image that closely resembles the original \cite{xia2022gan}. The core idea is to determine a latent vector $\boldsymbol{z}$ such that the generated image $G(z)$ is as similar as possible to the target image $X$. The inversion objective can be mathematically formulated as follows:
\begin{equation}z^*=\arg\min_z\mathcal{L}(G(z),X)\end{equation}
where $z^*\in Z$ represents the optimal latent vector in the latent space $Z$, and $G(\cdot)$ is the pre-trained generator function of the GAN that maps from $Z$ to the image space. The function $\mathcal{L}$ denotes a loss function that measures the discrepancy between the generated image $G(z)$ and the target image $X$. Commonly used loss functions include pixel-wise loss and perceptual loss. By minimizing this loss function, we aim to achieve a reconstructed image that closely approximates the original input image $X$. 

In our approach, we train an encoder to extract features from the input image. These features are then mapped into the latent space of the GAN for inversion. As shown in Fig. \ref{fig:2}, given an input image $X$, we use a pre-trained encoder to extract its features. The encoder produces two distinct outputs. The first output is the hierarchical latent code $w$, which controls the style attributes of each layer in the generator. The second output is an additional feature code $w^{*}$, which fine-tunes specific details during the generation process. These two codes are concatenated to form the latent space representation $Z_{R}$ in the GAN. This representation is then fed into a pre-trained StyleGAN to complete the GAN reconstruction, resulting in the reconstructed image $X_{RG}$.

\begin{figure}[t]
\centering
\includegraphics[width=0.98\linewidth]{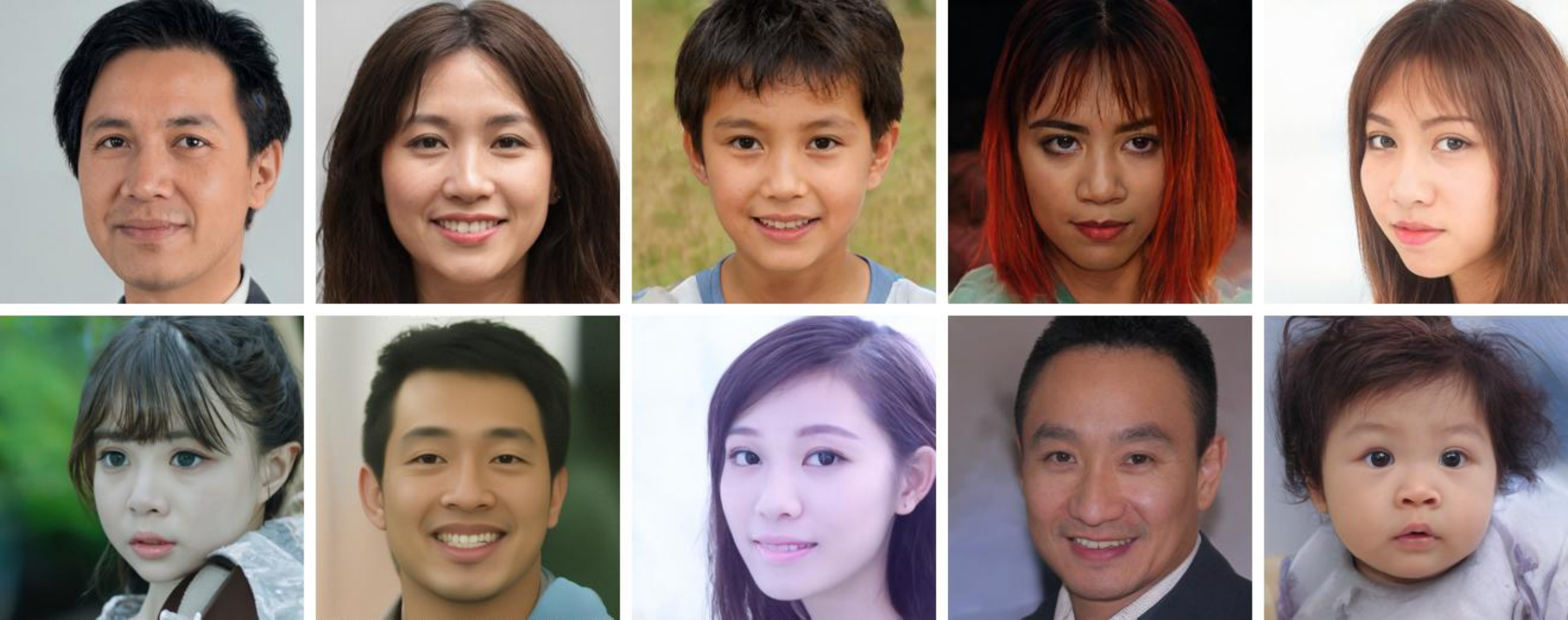}
\caption{Asian Synthetic Face Dataset. The first row contains images generated using different GANs, while the second row contains images generated using different DMs.}
\label{fig:1_b1}
\end{figure}

\begin{table*}[t]
    \renewcommand\arraystretch{0.9}
    \centering
    \footnotesize
    \caption{Comparison to Existing Detectors. Evaluation metrics for all experiments used Accuracy (ACC \%) and Average Precision (AP \%).}
    \label{tab:cross-mode}
    \begin{tabular}{c|ccccccccccccc|cc}  
        \toprule
    \multirow{4}{*}{Methods} & \multicolumn{13}{c|}{Testing Subset} & \multicolumn{2}{c}{\multirow{3}{*}{Avg}} \\ \cmidrule{2-14} 
    & Real & \multicolumn{2}{c}{StyleGAN1} & \multicolumn{2}{c}{ADM} & \multicolumn{2}{c}{StyleGAN2} & \multicolumn{2}{c}{IDDPM} & \multicolumn{2}{c}{VQGAN} & \multicolumn{2}{c|}{LDM}  \\ \cmidrule{2-14} 
    & ACC & ACC & AP & ACC & AP & ACC & AP & ACC & AP & ACC & AP & ACC & AP & ACC & AP \\ 
    \midrule
    CNNSpot \cite{wang2020cnn} & 99.5 & 100 & 100 & 99.8 & 99.9 & 36.4 & 92.6 & 68.5 & 86.4 & 1.4 & 45.7 & 18.1 & 80.3 & 60.5 & 84.2 \\
    LGRAD \cite{tan2023learning} & 0.2 & 99.5 & 98.3 & 98.2 & 47.4 & 10.2 & 44.9 & 88.6 & 26.4 & 9.8 & 45.5 & 89.1 & 26.6 & 56.5 & 48.2 \\
    DIRE \cite{wang2023dire} & 96.8 & 99.4 & 99.9 & 98.3 & 99.8 & 76.9 & 97.1 & 91.1 & 97.8 & 3.6 & 57.6 & 17.1 & 56.7 & 69.1 & 84.8 \\
    DeepFeatureX \cite{pontorno2024deepfeaturex} & 95.3 & 99.8 & 100 & 98.6 & 99.6 & 74.9 & 94.4 & 78.1 & 95.2 & 5.5 & 49.4 & 1.6 & 55.6 & 66.9 & 82.4 \\
    Cutting-Edge \cite{guarnera2024mastering} & 99.6 & 100 & 100 & 99.4 & 99.9 & 35.9 & 81.8 & 76.7 & 96.1 & 4.4 & 63.8 & 3.1 & 47.2 & 59.9 & 81.5 \\ 
    \midrule
    \textbf{Ours} & 99.9 & 99.8 & 100 & 99.3 & 99.9 & 91.4 & 99.9 & 97.7 & 99.5 & 25.5 & 64.1 & 56.1 & 73.7 & \textbf{81.6} & \textbf{89.5} \\ 
    \bottomrule
    \end{tabular}
\end{table*}

\subsubsection{DDIM Inversion and Reconstruction}
DMs \cite{sohl2015deep} generate data by progressive denoising through a series of inverse transformations, providing a structured approach to generative modeling. DDPM \cite{ho2020denoising} uses a Markov chain to iteratively denoise data, but this method is computationally expensive due to the large number of steps. To improve efficiency, DDIM \cite{song2020denoising} eliminates the Markov chain and offers a deterministic approach for both generation and inversion. In DDIM, the reverse process maps a noisy representation $x_{t}$ to the previous clean state $x_{t-1}$ deterministically, as shown in the equation:
\begin{equation}
\begin{split}
x_{t-1}=\frac{\sqrt{\alpha_{t-1}}x_t-\sqrt{1-\alpha_t}\epsilon_\theta(x_t,t)}{\sqrt{\alpha_t}}\\
+\sqrt{1-\alpha_{t-1}-\sigma_t^2}\cdot\epsilon_\theta(x_t,t)+\sigma_t\epsilon_t
\end{split}
\end{equation}
In this equation, $x_{t}$ represents the noisy image at time $t$, and $\epsilon_\theta(x_t, t)$ is the predicted noise from the model. As $\sigma_{t}$ approaches zero, the process becomes deterministic. This allows for approximate inversion from the clean image $x_{0}$ to the noisy representation $x_{T}$. This inversion process, known as \textbf{DDIM inversion}, is given by:
\begin{equation}x_{t+1}=\frac{x_t\sqrt{\alpha_t}}{\sqrt{\alpha_{t+1}}}+\sqrt{\frac{1-\alpha_{t+1}}{\alpha_{t+1}}-\frac{1-\alpha_t}{\alpha_t}}\cdot\epsilon_\theta(x_t,t)\end{equation}
While DDIM inversion is more efficient than DDPM, it still involves multiple steps. These steps can be computationally intensive. To address this issue, DDIM introduces a subset sampling strategy. It selects only a subset of $S$ steps $\{\tau_1,\ldots,\tau_S\}$, reducing computation while maintaining quality.

In our approach, we use the ADM \cite{dhariwal2021diffusion} network pre-trained on AFSD as the reconstruction model, and the DDIM \cite{song2020denoising} inversion and reconstruction process in which the number of steps $S=20$ by default. As shown in Fig. \ref{fig:2}, we perform DDIM inversion on the input image $X$, obtaining its latent representation $Z_T$. Then, we used a pre-trained ADM to reconstruct the image, resulting in the reconstructed image $X_{RD}$.

\subsubsection{Classification with Cascaded Images}
After obtaining the reconstructed images, unlike other reconstruction-based methods \cite{wang2023dire}, which calculate reconstruction residuals as classification inputs, our approach employs a different strategy. We argue that combining the original and reconstructed images as inputs provides richer feature representations, enhancing classification performance and model generalization. Consequently, as shown in Fig. \ref{fig:2}, the reconstructed images along with the original input images are cascaded and fed into a neural network for classification. In our implementation, we use ResNet-50 as the backbone. A ternary classifier is employed and optimized using ternary cross-entropy loss. The loss function is defined as: 
\begin{equation}L(y,\hat{y})=-\sum_{i=1}^Cy_i\log(\hat{y}_i)\end{equation}
Here, $y$ is the ground-truth one-hot encoded label vector, $\widehat{y}$ represents the predicted probability distribution over $C=3$ classes, and the sum is taken over all possible classes.

\subsection{Asian Synthetic Face Dataset}
Current synthesis face datasets prioritize racial diversity and broad coverage but lack dedicated datasets for Asian faces. Moreover, the quality of Asian face data is often lower, limiting the generalizability and effectiveness of existing models in applications specific to Asian populations. To address this, we introduce a new dataset, AFSD, as a complementary resource for Asian face datasets. Given the relatively uniform demographic characteristics of Asian populations, a synthetic face dataset tailored to Asian features could potentially provide researchers and developers with specialized resources to improve model performance in areas such as facial recognition, and privacy preservation, and serve as a benchmark for evaluating the effectiveness of synthetic image detectors.

The real images in the dataset are sourced from the FFHQ \cite{karras2019style} dataset, from which we selected 11,000 images specifically featuring Asian facial characteristics. We generated synthetic face images using four classical GAN models and four DM models. For GANs, we employed StyleGAN1 \cite{karras2019style}, StyleGAN2 \cite{karras2020analyzing}, ProGAN \cite{karras2018progressive}, and VQGAN \cite{esser2021taming}. The training code and pre-trained weights for these models are publicly available on GitHub. For DMs, we used ADM \cite{dhariwal2021diffusion}, IDDPM \cite{nichol2021improved}, LDM \cite{rombach2022high}, and SDE \cite{song2020score}. Each generative model produced 10,000 synthetic images at a resolution of 256x256 pixels. The dataset, which is depicted in Fig. \ref{fig:1_b1}, will be made publicly available to support research and community use. Furthermore, all experiments were conducted on our ASFD. The training set consisted of synthetic images generated by ADM and StyleGAN, along with the real images used to train them. The test set included images generated by StyleGAN2, VQGAN, IDDPM, and LDM, as well as the real images used for their training. Each type of image in the training set comprised 10,000 samples, while the test set contained 1,000 samples for each type.



\section{Experiment}

\subsection{Comparison to Existing Detectors }
To ensure fairness, all selected baselines were re-trained on our dataset using their publicly available training code. For the models in \cite{wang2020cnn}, \cite{tan2023learning}, and \cite{wang2023dire}, the classifier head was modified from binary to ternary classification. The quantitative results are presented in Table \ref{tab:cross-mode}.

Despite being re-trained on our dataset, most detectors exhibited a significant performance drop when handling the three-class classification task. Specifically, in detecting some unseen samples, the lowest ACC dropped below 20\%, and the lowest AP fell below 50\%. Models \cite{wang2020cnn}, \cite{guarnera2024mastering} and \cite{pontorno2024deepfeaturex} are data-driven approaches. As mentioned earlier, such methods struggle with more complex classification tasks. Although models \cite{guarnera2024mastering} and \cite{pontorno2024deepfeaturex} were designed for three-class classification, they fail to account for the differences between GAN and DM, resulting in poor generalization performance. Furthermore, the models proposed in \cite{tan2023learning} and \cite{wang2023dire} are detection networks specifically designed to identify GAN-generated images and DM-generated images, respectively. However, their performance diminishes in diverse and mixed scenarios, limiting their generalization capability. In particular, model \cite{tan2023learning} achieves an accuracy of only 0.2\% in detecting real images. This is because it extracts gradients using GANs for all input images. In the three-class classification task, gradients are obtained for GAN-generated, DM-generated, and real images using GANs. However, the gradient maps of DM-generated and real images are challenging to differentiate, causing the network to misclassify almost all real images as DM-generated. In contrast, our method achieved strong generalization, with an average ACC of 81.6\% and AP of 89.5\%, outperforming the best baseline by 12.6\% in ACC and 4.7\% in AP.

\begin{table}
\centering
\caption{Cross-Dataset evaluations.}
\label{tab:cross-datasets}
\begin{tabular}{ccc} 
\hline
Methods      & mean ACC & mean AP  \\ 
\hline
CNNSpot \cite{wang2020cnn}      & 34.98    & 78.58    \\
DIRE \cite{wang2023dire}         & 49.64    & 70.85    \\
Cutting-Edge \cite{guarnera2024mastering} & 37.92    & 74.96    \\
DeepFeatureX \cite{pontorno2024deepfeaturex} & 48.58    & 73.89    \\
Ours         & \textbf{62.91}    & \textbf{85.21}    \\
\hline
\end{tabular}
\end{table}

\subsection{Cross-Dataset Evaluations}
To further validate the effectiveness and generalization of our method, we conducted cross-dataset evaluations. Specifically, we selected synthetic face images from DiFF \cite{cheng2024diffusion} and several publicly available codes \cite{karras2019style,karras2020analyzing,rombach2022high} on GitHub and the real face images were sourced from FFHQ \cite{karras2019style}. Nearly all of these face images, both real and synthetic, feature Western individuals. Notably, we did not retrain the model but directly used the weights trained on ASFD, which provides a more stringent evaluation of the model's capability. The final test results are shown in Table \ref{tab:cross-datasets}. When faced with the new datasets, the ACC of all baseline models fell below 50\%, and their AP was significantly inferior to our model. Although our model had never encountered Western faces before, it analyzed the discrepancies between GANs and DMs from the perspectives of inversion and reconstruction. As a result, it achieved promising performance in these mixed scenarios.

\begin{table}
\centering
\caption{Ablation Evaluations.}
\label{tab:cross-model1}
\begin{tabular}{ccccc} 
\hline
Methods                                    & GAN & DM & mean ACC      & mean AP        \\ 
\hline
\multirow{3}{*}{Multi-Reconstruction} & $\checkmark$   &  -  & 72.8          & 82.9           \\
                                           &  -   & $\checkmark$  & 67.5          & 81.1           \\
                                           & $\checkmark$   & $\checkmark$  & \textbf{81.6} & \textbf{89.5}  \\ 
\hline
\multirow{3}{*}{Multi-Residual}       & $\checkmark$   &  -  & 70.9          & 84.1           \\
                                           &  -   & $\checkmark$  & 69.1          & 84.8           \\
                                           & $\checkmark$   & $\checkmark$  & \textbf{74.1} & \textbf{88.4} \\
\hline
\end{tabular}
\end{table}

\subsection{Multi- is crucial for generalization}
To show that the effectiveness of the Multi-Reconstruction-based Detector is not dependent on specific hyperparameter choices, we conducted ablation experiments using three input configurations: RGB images combined with GAN-reconstructed images, RGB images combined with DM-reconstructed images, and RGB images combined with multi-reconstructed images. As shown in Table \ref{tab:cross-model1}, the performance of models based on a single reconstruction was significantly lower than our proposed method. Specifically, feeding the input image along with its two reconstructions outperformed other configurations, highlighting the importance of multiple inversion and reconstruction for generalization.

Furthermore, following the suggestion from DIRE \cite{wang2023dire} that using residuals as inputs is more effective, we conducted experiments with different residual configurations. These configurations included residuals computed from GAN-reconstructed images $(Res = |X - X_{RG}|)$, residuals from DM-reconstructed images $(Res = |X - X_{RD}|)$ and multiple residuals from both reconstructions. As shown in Table \ref{tab:cross-model1}, the multiple residual approaches significantly outperformed single-residual configurations, further demonstrating the effectiveness of multiple inversion and reconstruction for generalization.

\begin{figure}[ht]
\centering
\includegraphics[width=1\linewidth]{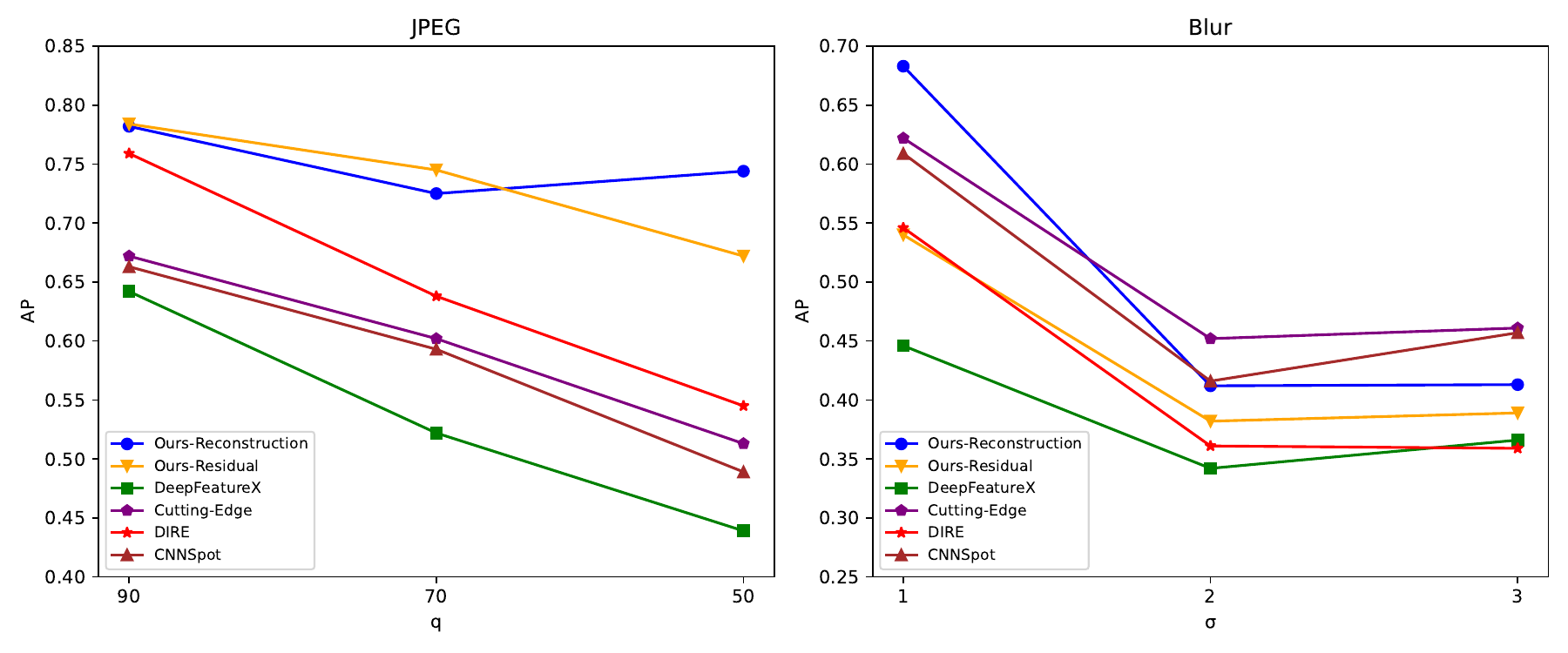}
\caption{Robustness evaluations against Gaussian blur and JPEG compression.}
\label{fig:rob}
\end{figure}

\subsection{Robustness to Unseen Perturbations }
In addition to evaluating generalization, robustness to unseen perturbations is a critical concern, as real-world images often undergo various distortions. Following the methodology of \cite{wang2020cnn}, we assessed the robustness of the detectors against two types of degradation: Gaussian blur and JPEG compression. Perturbations were applied at three levels of Gaussian blur $(\sigma=1, 2, 3)$ and three levels of JPEG compression $(q=1, 2, 3)$. As shown in Fig. \ref{fig:rob}, most methods performed poorly under perturbations, while our detector consistently demonstrated superior robustness.

\subsection{Qualitative Results and Visualizations }
\begin{figure}[t]
\centering
\includegraphics[width=1\linewidth]{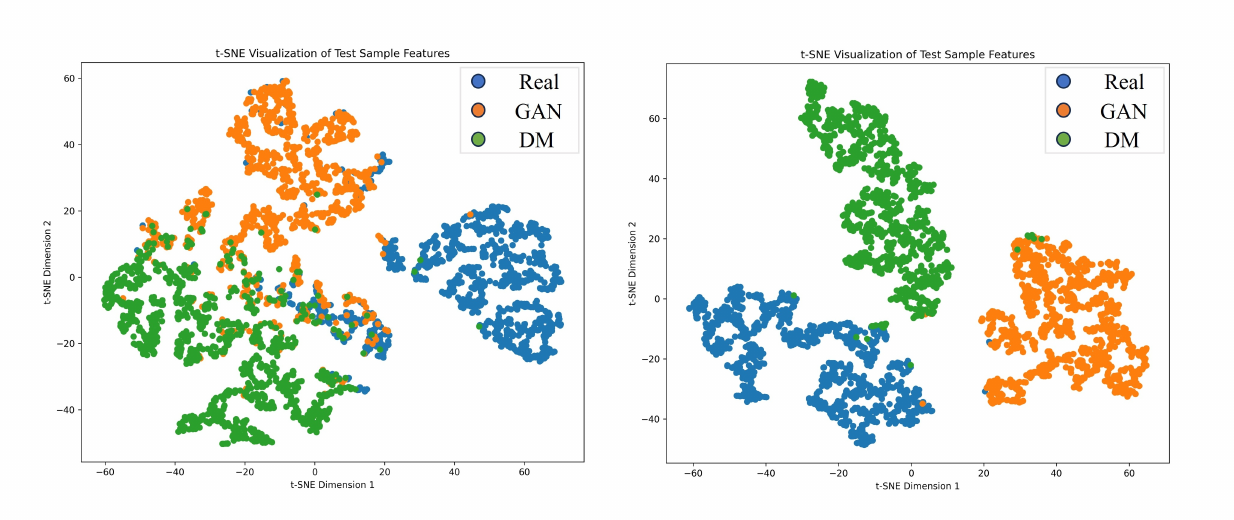}
\caption{Visualization of the spatial representation of the baseline model (left) and our model (right).}
\label{fig:spectrum}
\end{figure}

We use t-SNE \cite{van2008visualizing} visualization to illustrate the feature vectors extracted from the final layers of both our model and the baseline model \cite{guarnera2024mastering}, as shown in Fig. \ref{fig:spectrum}. The training of the model and experimental setup were consistent, with evaluations conducted on a subset of the ASFD. The baseline results (left) show significant overlap between GAN and DM images, while our model (right) effectively distinguishes real, GAN-generated, and DM-generated images with minimal overlap. This demonstrates our model's superior generalization capabilities.

\section{Conclusion}
In this paper, we perform discrepancy analysis to explore the intrinsic relationship between synthetic images and their generative techniques. We observe that reconstruction performance is superior when the reconstruction technique matches the generation method but degrades otherwise. Leveraging this observation, we propose a Multi-Reconstruction-based Detector that employs GAN and DM to invert and reconstruct input images. By capturing subtle discrepancies in reconstruction performance, our method effectively distinguishes between real, GAN-generated, and DM-generated images. Additionally, we curated a novel dataset, ASFD, to address the underrepresentation of Asian synthetic face data and provide crucial support and reference for tasks targeting Asian populations.

\section*{ACKNOWLEDGMENT}
This work was supported by the Open Research Fund of The State Key Laboratory of Blockchain and Data Security, Zhejiang University and supported by the National Natural Science Foundation of China Under Grant 62402182.

\bibliographystyle{IEEEtran}
\bibliography{main}

\end{document}